\documentclass[conference]{IEEEtran}
\IEEEoverridecommandlockouts

\usepackage{cite}
\usepackage{amsmath,amssymb,amsfonts}
\usepackage{algorithmic}
\usepackage{graphicx}
\usepackage{textcomp}
\usepackage{xcolor}
\usepackage{hyperref}
\usepackage{multirow}
\usepackage{subfigure}
\usepackage{balance}

\def\BibTeX{{\rm B\kern-.05em{\sc i\kern-.025em b}\kern-.08em
    T\kern-.1667em\lower.7ex\hbox{E}\kern-.125emX}}

\begin{document}

\title{Autonomous Racing using a Hybrid Imitation-Reinforcement Learning Architecture}

\author{Chinmay Vilas Samak$^{\ast}$, Tanmay Vilas Samak$^{\ast}$ and Sivanathan Kandhasamy%
\thanks{$^{\ast}$These two authors contributed equally to this research work. Chinmay Samak, Tanmay Samak and Sivanathan Kandhasamy are with Autonomous Systems Lab, Department of Mechatronics Engineering, SRM Institute of Science and Technology, Kattankulathur 603203, Tamil Nadu, India. {\tt\small {\{\href{mailto:cv4703@srmist.edu.in}{cv4703}, \href{mailto:tv4813@srmist.edu.in}{tv4813}, \href{mailto:sivanatk@srmist.edu.in}{sivanatk}\}@srmist.edu.in}}}%
}

\maketitle


\begin{abstract}
In this work, we present a rigorous end-to-end control strategy for autonomous vehicles aimed at minimizing lap times in a time attack racing event. We also introduce AutoRACE Simulator developed as a part of this research project, which was employed to simulate accurate vehicular and environmental dynamics along with realistic audio-visual effects. We adopted a hybrid imitation-reinforcement learning architecture and crafted a novel reward function to train a deep neural network policy to drive (using imitation learning) and race (using reinforcement learning) a car autonomously in less than 20 hours. Deployment results were reported as a direct comparison of 10 autonomous laps against 100 manual laps by 10 different human players. The autonomous agent not only exhibited superior performance by gaining 0.96 seconds over the best manual lap, but it also dominated the human players by 1.46 seconds with regard to the mean lap time. This dominance could be justified in terms of better trajectory optimization and lower reaction time of the autonomous agent.
\end{abstract}



\section{Introduction}
\label{section: Introduction}

The task of autonomous driving has been researched for several years now, and although engineers have been able to demonstrate successful deployment of self-driving cars in the real-world, their reliability is still questionable. Accidents caused by these vehicles over public roads is another area of concern, wherein the responsibility of the act is a highly fragile issue. However, the issue of ethical clearance is significantly reduced if the same technology is given a slightly different application -- that of racing. Autonomous racing in confined and reinforced racetracks is increasingly gaining popularity as a new form of sport/entertainment, since these vehicles not only showcase extreme engineering, but also exhibit excellent acrobatics. This makes it an interesting problem statement to address, wherein the aim is to stretch the limits of an autonomous agent to race around a closed circuit, possibly even surpassing human performance.

However, autonomous racing is an elevated engineering problem requiring extremely timed rigorous maneuvers. This calls for minimizing latency of the autonomy software, which is affected by the overall system architecture. Additionally, it demands highly efficient trajectory planning and control to track what is termed as the \textit{racing line}, which allows a racer to drive in a straighter line before reaching frictional limits, thereby cornering turns in a much aggressive manner. A typical racing line is defined following an \textit{out-in-out} maneuver, wherein the vehicle drives form outer side of the track up to the \textit{turn-in point}, then turns in while aiming for the \textit{geometric apex} of the curve, and again runs wide after the turn. Nevertheless, mastering a racing line requires extreme focus and years' worth of practice behind the wheel; but once done, expert racers are able to minimize drifting and maneuver their vehicles perfectly based on the steering commands alone.

\begin{figure}[t]
	\centering
	\includegraphics[width=\linewidth]{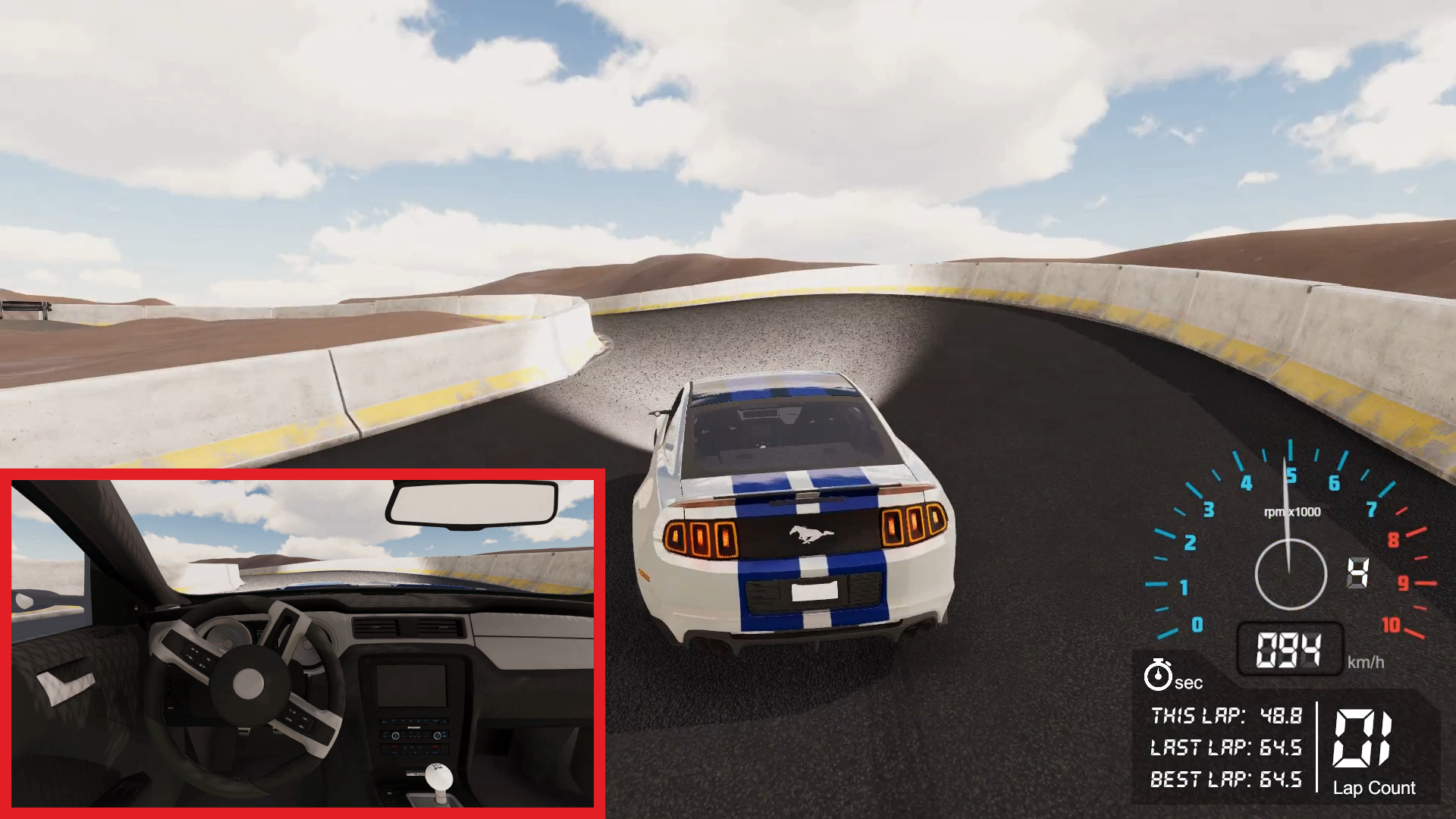}
	\caption{Still from the fastest lap of virtual time attack car racing event within AutoRACE Simulator depicting a ``Need for Speed'' 2013 Ford Mustang GT being controlled by our autonomous agent trained using a hybrid learning approach. Notice the FPV and HUD at the bottom-left and bottom-right corners, respectively.}
	\label{fig1}
\end{figure}

This work introduces AutoRACE Simulator (refer Fig. \ref{fig1}), an open, flexible and realistic dynamic simulation system specifically targeted towards autonomous racing. We employ the said simulator to train a machine learning (ML) agent to race autonomously in an end-to-end manner. Particularly, we adopt a hybrid learning strategy along with a novel reward function, wherein the autonomous agent implicitly learns trajectory optimization to ultimately minimize lap times in a time attack car racing event. We show that our autonomous agent is able to outperform human players while exhibiting significant dominance over them.

This work can be applied to virtual racing games in order to develop realistic non-player characters (NPCs) that compete with human players solely based on their superior trajectory optimization and lower reaction time, rather than any unfair means. Additionally, owing to its simplicity, this approach can also be transferred to physical race cars to compete in real-world autonomous and/or mixed racing events.


\section{Related Work}
\label{section: Related Work}

Autonomous car racing is a relatively new field undergoing constant research and development. The prior works in this field can be categorized based on their principal approach of motion control as either \textit{modular} or \textit{end-to-end}.

\subsection{Modular Approaches}
\label{subsection: Modular Approaches}

Modular approaches address the task of autonomous driving by dividing it into three fundamental sub-systems viz., perception, planning and control. Simpler approaches target the perception and planning sub-systems, while implementing error-driven controllers such as PID \cite{marino2009, guvenc2014} or geometric path tracking controllers such as Pure-Pursuit \cite{coulter1992} or Stanley \cite{hoffmann2007}. On the other hand, advanced approaches are mostly concerned with planning and tracking an optimal trajectory. Particularly, the task of time-optimal trajectory planning (TOTP) can be addressed in a number of ways: \cite{velenis2005} maximizes the exit velocity of the vehicle during cornering, \cite{timings2013} maximizes the distance traveled by the vehicle along the track centerline in a fixed time interval, and \cite{rucco2015} directly minimizes the travel time of the vehicle subject to control and frictional limits. Model predictive control (MPC) \cite{liniger2015, williams2016, alcala2020} is another widely adopted technique to generate and track an optimal trajectory following the receding horizon principle, which assumes an accurate vehicle model and near-perfect state estimation. Furthermore, some approaches \cite{williams2017, kabzan2019} have also tried adopting deep neural networks as non-linear function approximators to learn/improve the vehicle models as a way of addressing the underlying imperfections and uncertainties.

Although these approaches are more predictable in terms of their performance and can be trusted for the same reason, they suffer from a higher chance of failure; fault(s) within any sub-system or miscommunication among the sub-systems potentially leads to the failure of the entire system. Additionally, most of the advanced approaches in this category have a significantly high computational overload considering embedded applications, especially for perception and planning tasks, and optimal controllers only add to this problem.

\subsection{End-to-End Approaches}
\label{subsection: End-to-End Approaches}

End-to-end approaches exploit the notion of directly mapping the sensory data to actuation command(s) by abstracting the perception, planning and control functions into a deep neural network model. Not only does this greatly reduce the software latency, but it also eliminates the need of having a priori knowledge about the vehicle or its environment.

Imitation learning relies on a labeled dataset to learn the task of autonomous driving by training a deep neural network. This approach was first implemented by \cite{pomerleau1989} in order to predict direction of travel based on camera and laser range finder feed, using a fully connected neural network (FCNN). Recently, \cite{muller2006} adopted a convolutional neural network (CNN) to learn off-road obstacle avoidance behavior based on stereo camera feed. Building on top of the prior works, \cite{bojarski2016, bojarski2017} trained a CNN model to steer an autonomous vehicle by implicitly detecting the road lanes and analyzed its feature extraction capabilities. More recently, \cite{samak2020} investigated end-to-end imitation learning more exhaustively and reported performance metrics to evaluate robustness of models trained using this approach. However, in terms of autonomous racing, imitation learning approaches tend to be sub-optimal since they heavily rely on the labeled dataset recorded by an expert, and are therefore upper-bounded by the quality of training data. In simpler terms, they may not surpass the performance of expert that recorded the dataset.

Reinforcement learning optimizes parameterized policies based on reward signals acquired through self-exploration. It does not depend on a labeled dataset, which allows an agent trained using this approach to potentially surpass human expertise. However, this approach, in its pure form, is extremely time consuming; not to mention the risk of \textit{reward hacking}, which leads to agents learning potentially unwanted behaviors. Prior works \cite{riedmiller2007, kendall2019, jaritz2018} have employed reinforcement learning for autonomous driving without the absolute intention of racing. Some of the recent works \cite{fuchs2020, song2021} have also applied this technique to address the problem of autonomous racing, but they made use of extensively large (and somewhat unrealistic) observation space with explicit feature extraction, which may be only limited to virtual gaming applications.

This work builds on the conclusions and shortcomings derived from previous works to demonstrate super-human autonomous racing in a simulated setting. We adopt a hybrid imitation-reinforcement learning architecture, and make use of practically feasible observation and action spaces in order to enable \textit{sim2real} transferability in the future. We show that our approach significantly reduces unwanted exploration and reward hacking, which ensures both faster and safer training \cite{samak2021}, even with limited computational resources.


\section{Methodology}
\label{section: Methodology}

While the primary objective of this research project was to leverage an end-to-end imitation-reinforcement learning strategy to efficiently train an autonomous agent to race aggressively, implementing and validating our approach required the development of a dynamic simulation system allowing realistic interaction between the agent and the environment, while offering the flexibility to work with a hybrid learning architecture. We first discuss the simulation system and its features before delving deeper into the specifics of the learning architecture and experimentation adopted for this work.

\subsection{Simulation System}
\label{subsection: Simulation System}

The present research community lacks dedicated simulation tools to virtually prototype autonomous racing algorithms. Particularly, currently available open-source simulators such as Gazebo \cite{koenig2004} and TORCS \cite{espi2005} fail to capture realistic system dynamics and graphics required for effective sim2real transfer. Conversely, commercial automotive simulators and video games such as Gran Turismo Sport \cite{gts} offer stunning realism, but are protected by end-user license agreements, which renders them moot for implementing novel approaches or reproducing and validating the results of a prior work, unless and otherwise special permissions allowing such activities are acquired. Moreover, in addition to licensing policies, most of the high-end simulators and video games demand additional hardware (consoles, controllers, networking equipment, etc.) and software (custom application programming interfaces, addon packages, etc.) resources to run and interact with them, especially for developing autonomy solutions.

AutoRACE Simulator\footnote{\textbf{Source code:} \href{https://github.com/Tinker-Twins/AutoRACE-Simulator}{https://github.com/Tinker-Twins/AutoRACE-Simulator}} bridges this quality-flexibility gap by providing an open-source, high-fidelity simulation ecosystem aimed at accelerated development, deployment and analysis of autonomous racing algorithms. The simulator is developed atop the Unity \cite{unity} game engine and has been integrated with ML-Agents Toolkit \cite{juliani2020} in order to allow flexible implementation of both modular and end-to-end algorithms. It exploits NVIDIA's PhysX engine to simulate detailed vehicular dynamics and accurate environmental physics at the backend, and employs Unity's high-definition rendering pipeline (HDRP) and post-processing stack to render photorealistic graphics on the frontend.

\begin{figure}[t]
	\centering
	\includegraphics[width=\linewidth]{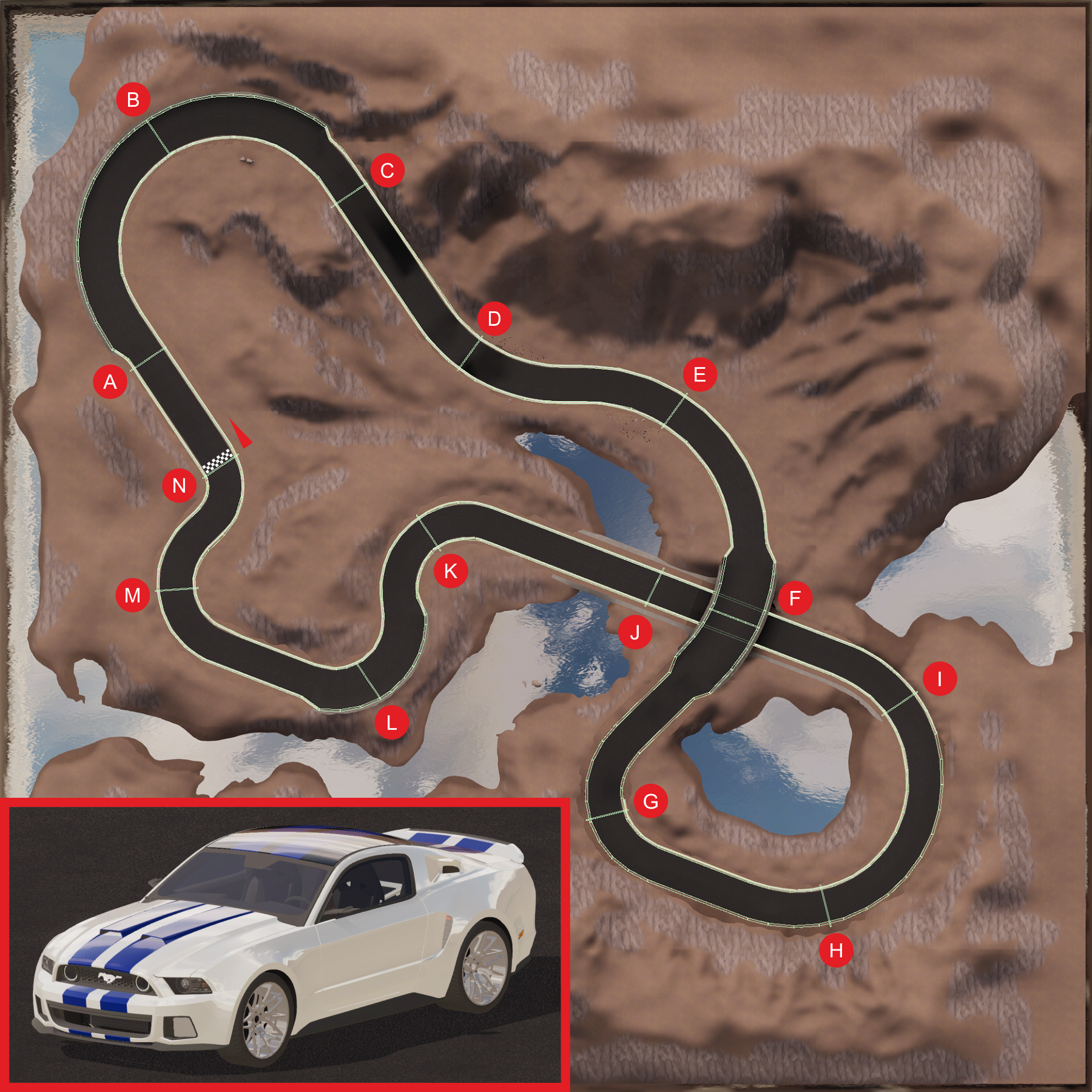}
	\caption{Simulation setting comprising a 500 $\times$ 500 m ``Oasis'' environment with racetrack, roadside barriers and 14 virtual checkpoints $\left \{ \text{A}, \text{B}, \cdots, \text{N} \right \}$. Notice the simulated ``Need for Speed'' 2013 Ford Mustang GT model at the bottom-left corner.}
	\label{fig2}
\end{figure}

Although it has been stressed multiple times that AutoRACE Simulator is targeted towards autonomous racing applications, it also supports manual driving by making use of a heuristic controller, which overrides the autonomous agent. This can be particularly useful for comparative performance analysis of the autonomous agent against human experts. The simulator currently features a short conceptual racetrack as the racing environment, and a ``Need for Speed'' 2013 Ford Mustang GT model as the simulated vehicle (refer Fig. \ref{fig2}).

The primary justification governing the vehicle choice is the versatility of Ford Mustang GT to be employed for almost all types of races, owing to its high power delivery, speed and agility. The simulated vehicle resembles its real-world counterpart in terms of aesthetics as well as dynamics. It features realistic control inputs (throttle, brake, handbrake, steering) along with a 6-speed automatic transmission, while offering the flexibility of controlling any one (or more) of these operations automatically through backend scripting. It is programmed with detailed dynamic models for the engine, drive-train, suspensions and tires, which are tuned close to those of the real vehicle. Furthermore, physical effects such as air drag, traction, friction, drifting, cornering, collision and momentum transfer have also been modeled accurately in order to close the gap between simulation and reality.

The simulated racing environment is given the appearance of an oasis, and the racetrack is designed to be self-sufficient in terms of driving features (refer Fig. \ref{fig2}): it comprises 3 large curves (ABC, DEF, GHI), 2 short deviations (CDE, EFG), 5 straights (NA, CD, GH, IJK, LM), 4 sharp turns (JKL, KLM, LMN, MNA), as well as an overpass (EFG) and underpass (IJ). The length of the track is kept short so as to conform with the Fédération Internationale de l'Automobile (FIA) standards of measuring its length in terms of laps instead of the actual driving distance, thereby supporting time attack racing (employed in this work). The physical attributes of the environment have been modelled with the help of baked lighting, rigid body components, mesh colliders and physical materials to accurately simulate the interaction between the vehicle and the environment. Another key addition to the racing environment is that of 14 virtual checkpoints (A-N), illustrated in Fig. \ref{fig2}. These checkpoints essentially guide the autonomous agent to progress through the lap by providing appropriate incentive upon passing each checkpoint. The final checkpoint (N) is effectively the finish line, crossing which, completes a lap.

In addition to simulating vehicular dynamics and environmental physics, the simulator also offers realistic audiovisuals including wheel effects such as skid marks and tire smoke, headlights and taillights with accurate raycasting and diffusion, audios for engine and tire sounds, animations for vehicle wheels and steering wheel, etc. Moreover, it also supports two distinct camera views (refer Fig. \ref{fig1}): one that follows the vehicle from behind, giving a third-person view (TPV), and other that sits in the driver's seat, giving an excellent first-person view (FPV). It is worth noting that the camera motions are updated dynamically to give a sense of inertial effect to the human drivers. Furthermore, the simulator also features an automated data logging functionality to record racing events for further analysis. Additionally, users can analyze vital system parameters and lap data on the heads-up display (HUD), depicted in Fig. \ref{fig1}.

AutoRACE Simulator natively supports C\# scripting but we plan to develop and release APIs for C++, Python and ROS in the near future. Other potential updates include new vehicle models, scenarios, race settings and multiplayer support.

\begin{figure*}[t]
	\centering
	\includegraphics[width=\linewidth]{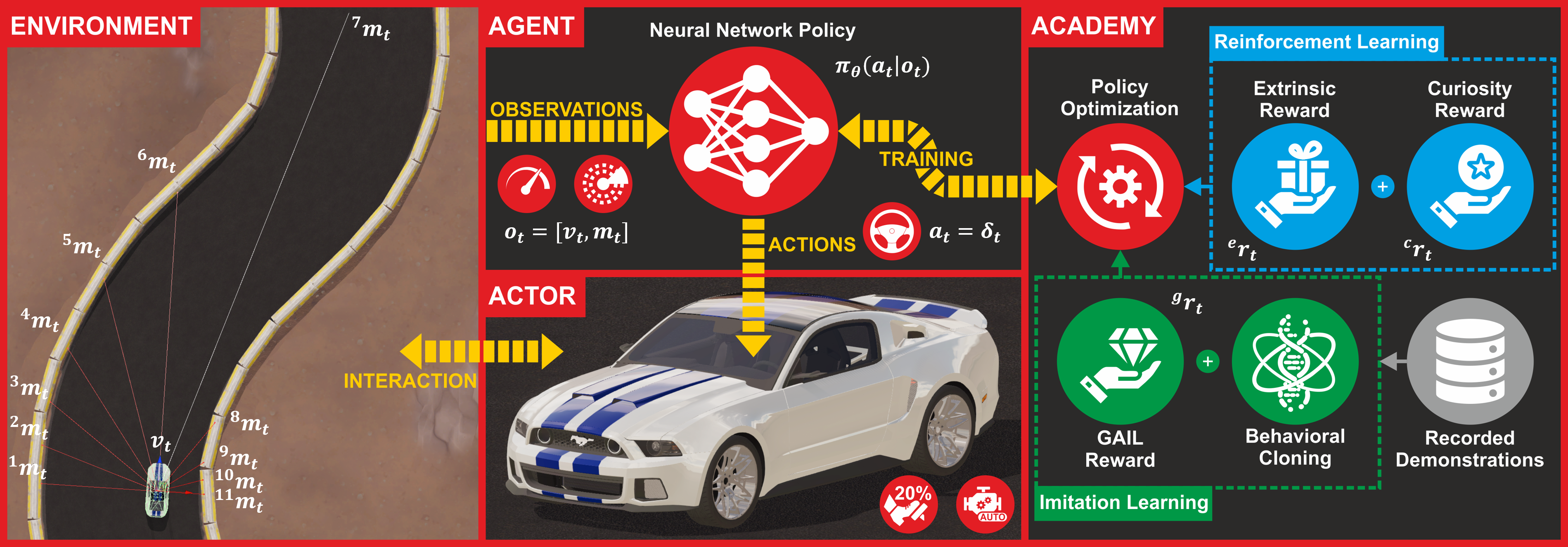}
	\caption{Hybrid imitation-reinforcement learning architecture adopted to train a deep neural network policy to race autonomously in an end-to-end manner. Note the clear distinction and interaction among various elements of the architecture: environment, agent, actor and academy.}
	\label{fig3}
\end{figure*}

\subsection{Learning Architecture}
\label{subsection: Learning Architecture}

The proposed learning architecture is a hybrid of imitation and reinforcement learning techniques, and is composed of the following key elements:

\begin{itemize}
	\item \textbf{Environment:} Environment comprises active and passive elements that may physically interact with the actor(s). These elements can be observed and acted upon.
	\item \textbf{Agent:} Agent is an intelligent entity, which can observe its environment and learn to predict its actions based on those observations.
	\item \textbf{Actor:} Actor is a physical entity, which interacts with the environment based on the decisions of the agent controlling it. The terms \textit{actor} and \textit{vehicle} are interchangeably used in this context.
	\item \textbf{Academy:} Academy orchestrates training and decision making process of the agent(s).
\end{itemize}

Initially, we recorded a demonstration dataset by letting a human drive the vehicle around the racetrack for 11 laps (using a standard keyboard). These demonstrations were sub-optimal in terms of trajectory planning and were solely intended to impart fundamental driving ability to the autonomous agent. We then adopted the hybrid learning architecture illustrated in Fig. \ref{fig3} to impart autonomous racing ability to the agent. It is interesting to note that while reinforcement learning recursively improved the agent's racing performance, imitation learning ensured that the agent equally focused on its driving ability during exploration phase, which minimized undesirable exploration (such as trying actions leading to a crash), thereby significantly reducing the training time. Table \ref{tab1} describes the training configuration adopted for this work.

\begin{table}[t]
	\caption{Training Configuration}
	\begin{center}
		\begin{tabular}{l|l}
			\hline
			\textbf{HYPERPARAMETER}                   & \textbf{VALUE}        \\ \hline
			\multicolumn{2}{l}{\textbf{Common Parameters}}                               \\ \hline
			Neural network architecture (FCNN)        & 2 $\times$ \{128, Swish \cite{ramachandran2017}\} \\
			Batch size                                & 64                    \\
			Buffer size                               & 1024                  \\
			Learning rate ($\alpha$)                  & 3e-4                  \\
			Learning rate schedule                    & Linear                \\
			Entropy regularization strength ($\beta$) & 0.01                  \\
			Policy update hyperparameter ($\epsilon$) & 0.2                   \\
			Regularization parameter ($\lambda$)      & 0.97                  \\
			Epochs                                    & 3                     \\
			Maximum steps                             & 5e6                   \\ \hline
			\multicolumn{2}{l}{\textbf{Behavioral Cloning}}                   \\ \hline
			Strength                                  & 0.5                   \\ \hline
			\multicolumn{2}{l}{\textbf{GAIL Reward}}                          \\ \hline
			Discount factor ($^{g}\gamma$)            & 0.99                  \\
			Strength                                  & 0.01                  \\
			Encoding size                             & 128                   \\
			Learning rate ($^{g}\alpha$)              & 3e-4                  \\ \hline
			\multicolumn{2}{l}{\textbf{Curiosity Reward}}                     \\ \hline
			Discount factor ($^{c}\gamma$)            & 0.99                  \\
			Strength                                  & 0.02                  \\
			Encoding size                             & 256                   \\
			Learning rate ($^{c}\alpha$)              & 3e-4                  \\ \hline
			\multicolumn{2}{l}{\textbf{Extrinsic Reward}}                     \\ \hline
			Discount factor ($^{e}\gamma$)            & 0.99                  \\
			Strength                                  & 1.0                   \\ \hline
		\end{tabular}
	\end{center}
	\label{tab1}
\end{table}

At each time step $t$, the agent collected a vectorized observation $o_t$ \eqref{eqn1} from the environment using velocity estimation and exteroceptive ranging modalities mounted on the virtual vehicle:
\begin{equation}
o_t = \left [ v_t, m_t \right ] \in \mathbb{R}^{12}
\label{eqn1}
\end{equation}
where, $v_t \in \mathbb{R}^{1}$ is the forward velocity of the vehicle within the environment and $m_t = \left [ ^{1}m_t, ^{2}m_t, \cdots, ^{11}m_t \right ] \in \mathbb{R}^{11}$ is the measurement vector providing 11 frontal range readings up to 50 meters, uniformly distributed 90$^{\circ}$ around each side of the heading vector, equally spaced 18$^{\circ}$ apart (refer Fig. \ref{fig3}). These observations were fed to a deep neural network policy $\pi_{\theta}$, wherein the weights and biases of the neural network formed the policy parameters $\theta \in \mathbb{R}^d$. The policy then directly mapped the observations $o_t$ to an appropriate action $a_t$ \eqref{eqn2} based on its experience so far:
\begin{equation}
a_t = \delta_t \in \mathbb{R}^{1}
\label{eqn2}
\end{equation}
where, $\delta_t \in \left \{ -1, 0, 1 \right \}$ is the discrete steering command indicating left, straight and right turn respectively. For longitudinal control, the throttle $\tau_t$ was set to 20\% of its upper saturation limit, which, in conjunction with the automatic transmission, determined the vehicle's forward velocity. These control commands $u_t = \left [ \delta_t, \tau_t \right ]$ were transmitted to the virtual actuators of the vehicle in order to control it autonomously. Meanwhile, the policy $\pi_{\theta}$ was optimized based on the following signals:

\begin{itemize}
	\item \textbf{Behavioral Cloning:} This was the core imitation learning algorithm, which updated the policy in a supervised fashion with respect to the recorded demonstrations. The behavioral cloning \cite{bain1995} update was carried out every once in a while, mutually exclusive of the reinforcement learning update.
	\item \textbf{GAIL Reward:} The generative adversarial imitation learning (GAIL) reward \cite{ho2016} $^{g}r_t$ ensured that the agent optimized its actions in a safe and ethical manner by rewarding proportional to the closeness of new observation-action pairs to those from the recorded demonstrations. This allowed the agent to retain its autonomous driving ability throughout the training process.
	\item \textbf{Curiosity Reward:} The curiosity reward \cite{pathak2017} $^{c}r_t$ promoted exploration by rewarding proportional to the difference between predicted and actual encoded observations. This ensured that the agent effectively explored its environment, even though the extrinsic reward was sparse.
	\item \textbf{Extrinsic Reward:} In the context of reinforcement learning, the objective of lap time reduction and motion constraints were handled using a novel extrinsic reward function $^{e}r_t$ \eqref{eqn3}, which guided the agent towards optimal behavior using the proximal policy optimization (PPO) \cite{schulman2017} algorithm. The agent was awarded $r_{checkpoint}=+1$  for crossing each of the checkpoints $c_i; \: i \in \left [ \text{A}, \text{B}, \cdots, \text{N} \right ]$ on the racetrack (refer Fig. \ref{fig2}), $r_{best\:lap}=+10$ upon beating the best lap time so far, and was penalized by $r_{collision}=-100$ for colliding with any of the roadside barriers $b_j; \: j \in \mathbb{R}^{n}$. Apart from these event-based rewards/penalties, the agent was continuously rewarded proportional to its velocity $v_t$, which ultimately forced it to optimize its trajectory spatio-temporally.
	\begin{equation}
	^{e}r_t  =  
	\begin{cases}
	r_{collision} & \text{if collided} \\
	r_{checkpoint} & \text{if passed checkpoint} \\
	r_{best\:lap} & \text{if recorded best lap time} \\
	0.01*v_t & \text{otherwise}
	\end{cases}
	\label{eqn3}
	\end{equation}
\end{itemize}

\subsection{Experimentation}
\label{subsection: Experimentation}

The trained autonomous agent was subjected to a time attack racing experiment, wherein its performance was compared against human experts. However, this required certain measures to be taken in order to reach an unbiased decision.

First of all, a direct comparison of the autonomous agent against human players required sufficient data to be collected. For this purpose, we recorded and compared a total of 10 autonomous laps against 100 manual laps by 10 different human players (i.e., 10 laps per player). Particularly, the autonomous racing performance was recorded by incorporating the trained neural network model into the vehicle’s control architecture and executing the developed pipeline in inference mode for 10 consecutive laps. On the other hand, manual racing performance was recorded by allowing 10 test subjects, experienced in the art of virtual car racing, to control the vehicle in heuristic mode and generating a log file for 10 complete laps. It is to be noted, however, that all the human players were allowed to practice 10 laps around the circuit in order to get acquainted with the simulator, after which, their racing performance was recorded.

Secondly, the playing field was to be leveled so as to restrict the autonomous agent as well as human players from practicing any unfair advantages over each other. This was achieved as follows:

\begin{itemize}
	\item \textbf{Similar Approach:} This work adopted an end-to-end approach, which is similar to decision making process of humans. This ensured that both the entities had a similar learning/deployment architecture, wherein they had to predict the control actions directly based on the real-time observations.
	\item \textbf{Equivalent Observations:} The observation vector for the autonomous agent comprised of velocity and ranging measurements, which are easily interpretable for a machine. On the other hand, human players had to estimate the pose and velocity of the vehicle based on visual feedback from the simulator, which is the most comprehensible form of perception for humans. Nevertheless, human players were free to choose between TPV and FPV cameras as per their convenience, and view the HUD if required.
	\item \textbf{Appropriate Actions:} The longitudinal control in either case was implemented with a constant throttle, which ensured that both the entities had to rely on trajectory optimization and control precision in order to ultimately reduce their respective lap times. Additionally, since the human players were to use a standard computer keyboard to control the vehicle, the autonomous agent was also made to choose its actions from a discrete domain.
\end{itemize}


\section{Results}
\label{section: Results}

We report our findings in terms of quantitative figures and qualitative remarks for the training and deployment phases, followed by a discussion on the computational details of our approach.

\subsection{Training Analysis}
\label{subsection: Training Analysis}

\begin{figure}[t]
	\centering
	\subfigure[]{\includegraphics[width=0.24\textwidth]{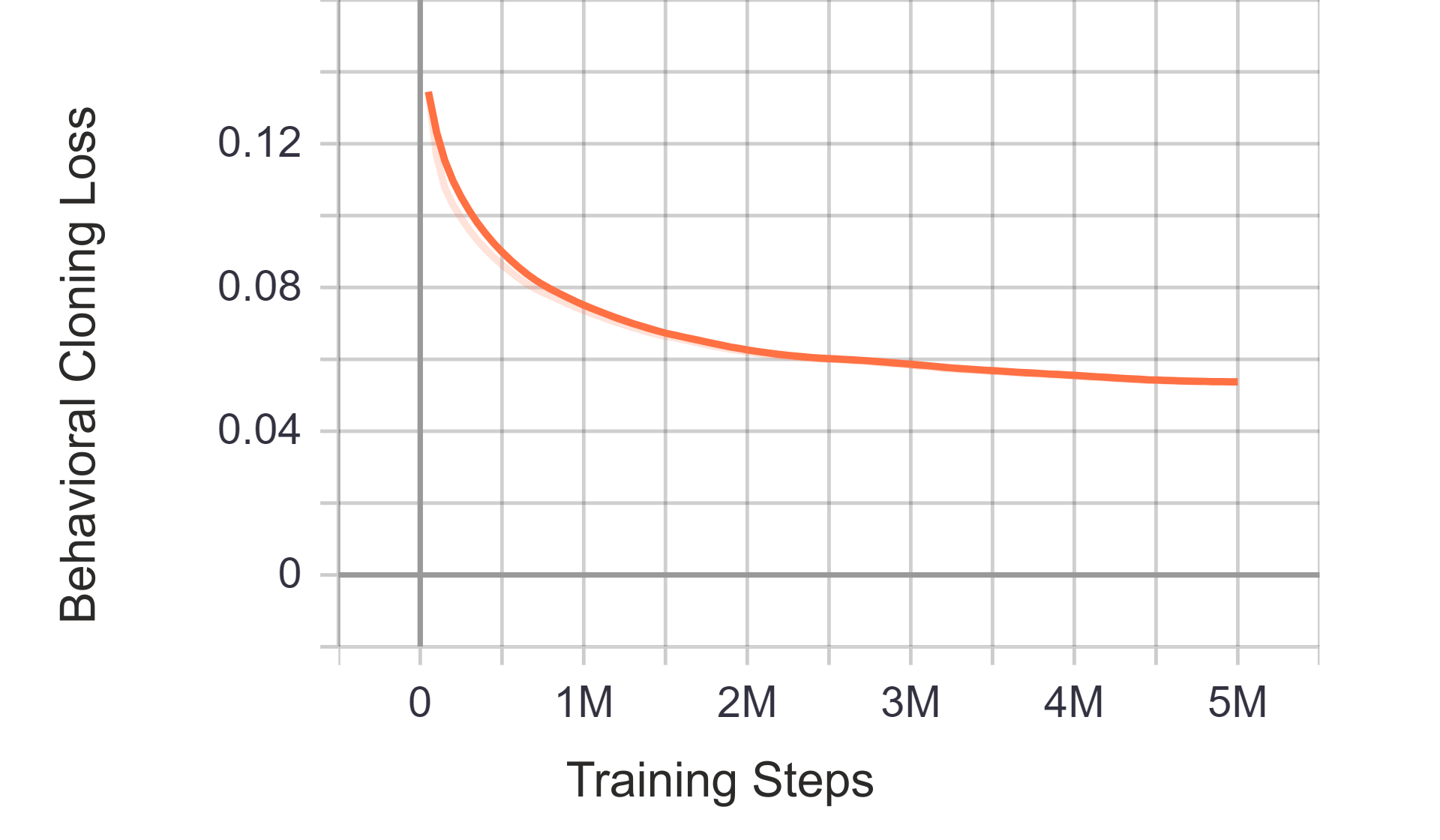}}
	\subfigure[]{\includegraphics[width=0.24\textwidth]{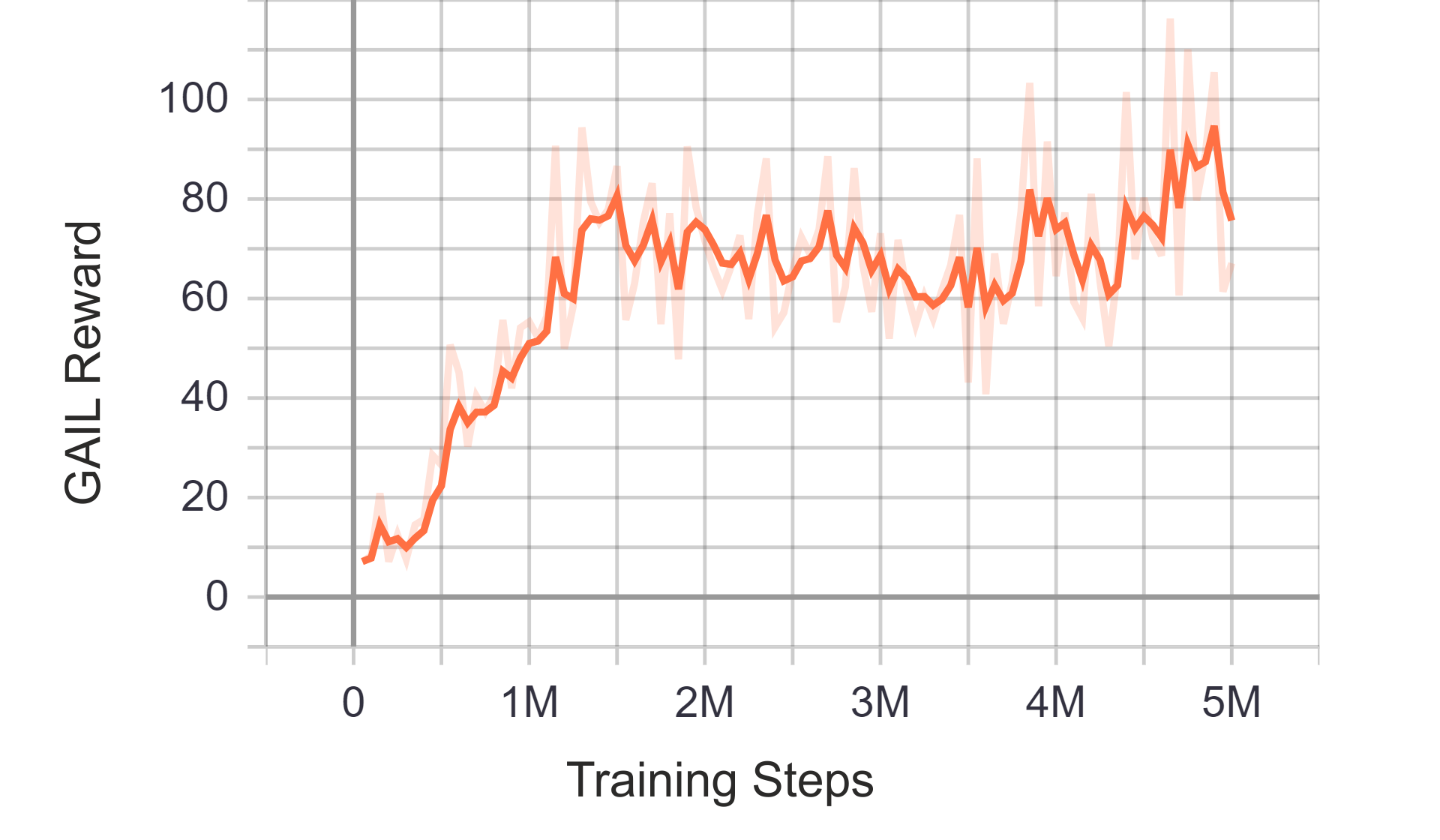}}
	\subfigure[]{\includegraphics[width=0.24\textwidth]{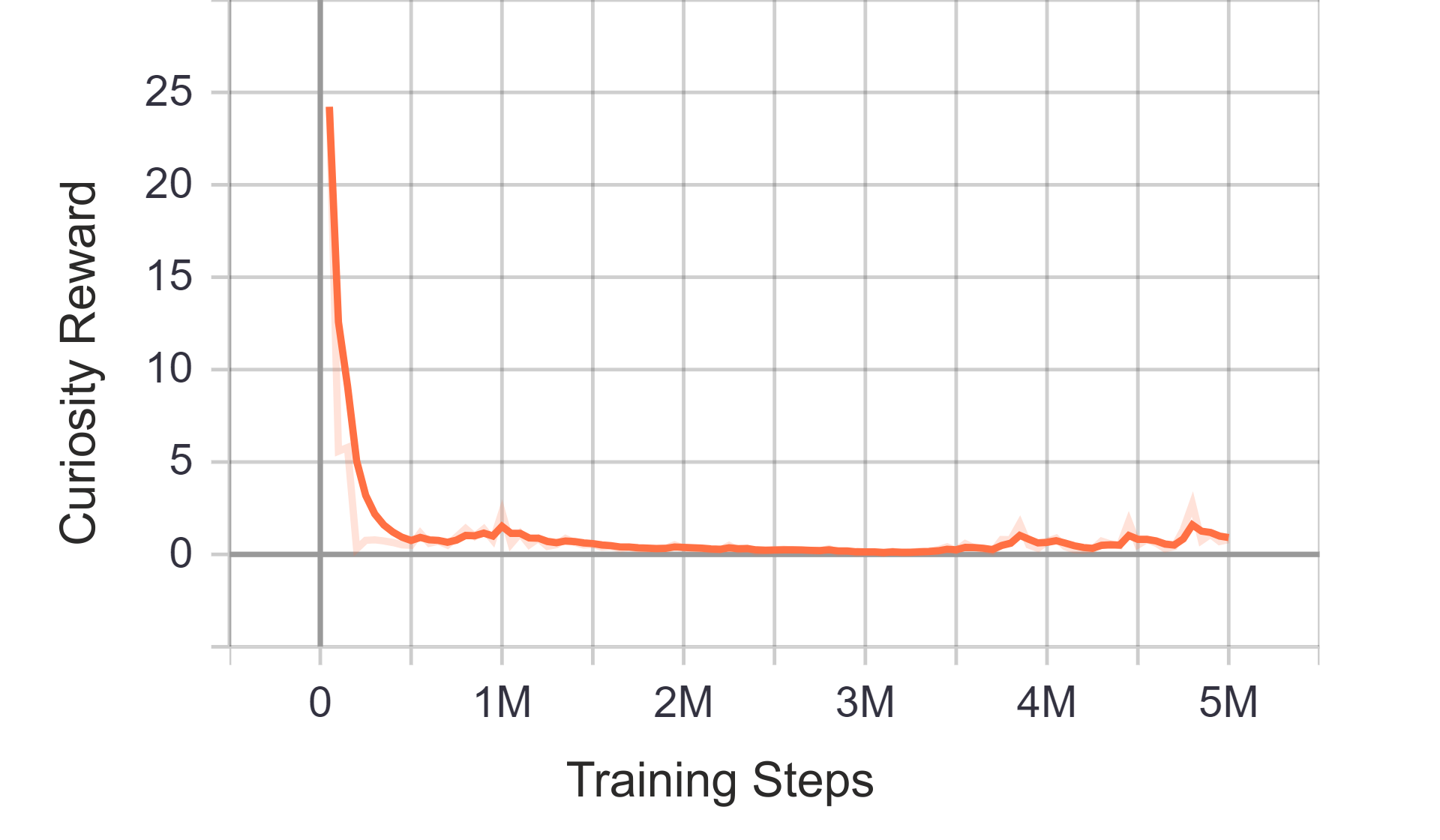}}
	\subfigure[]{\includegraphics[width=0.24\textwidth]{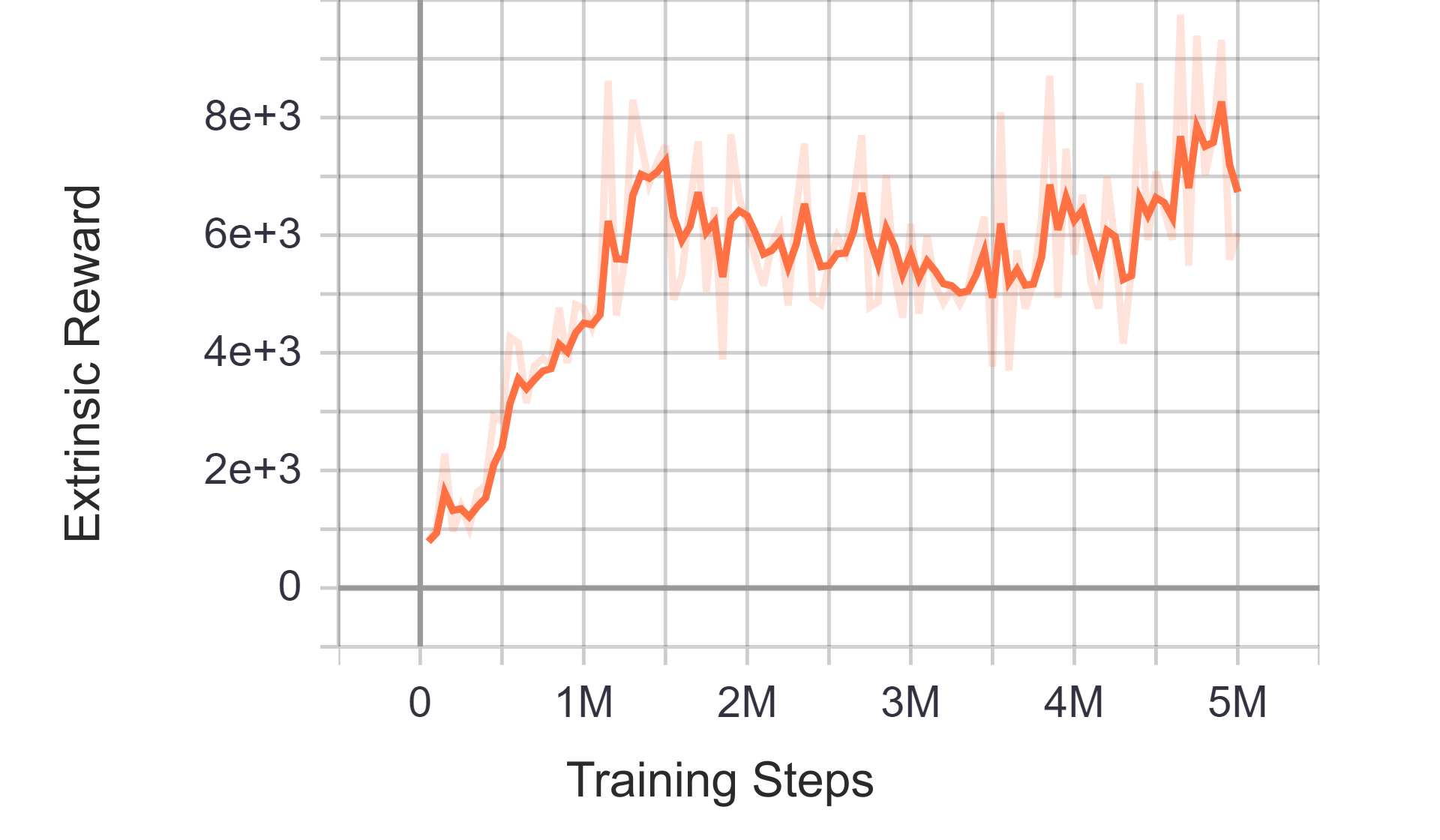}}
	\subfigure[]{\includegraphics[width=0.24\textwidth]{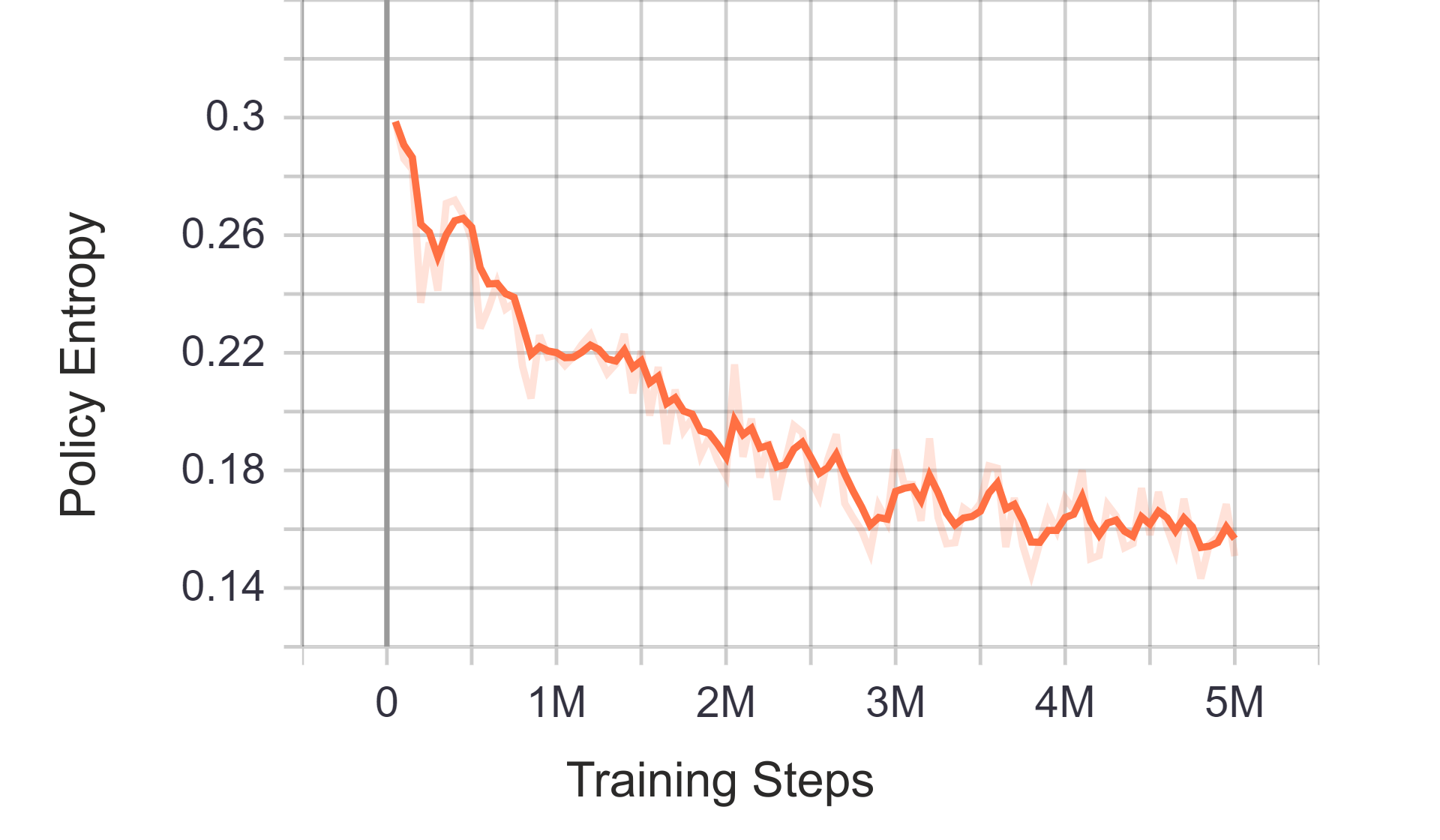}}
	\subfigure[]{\includegraphics[width=0.24\textwidth]{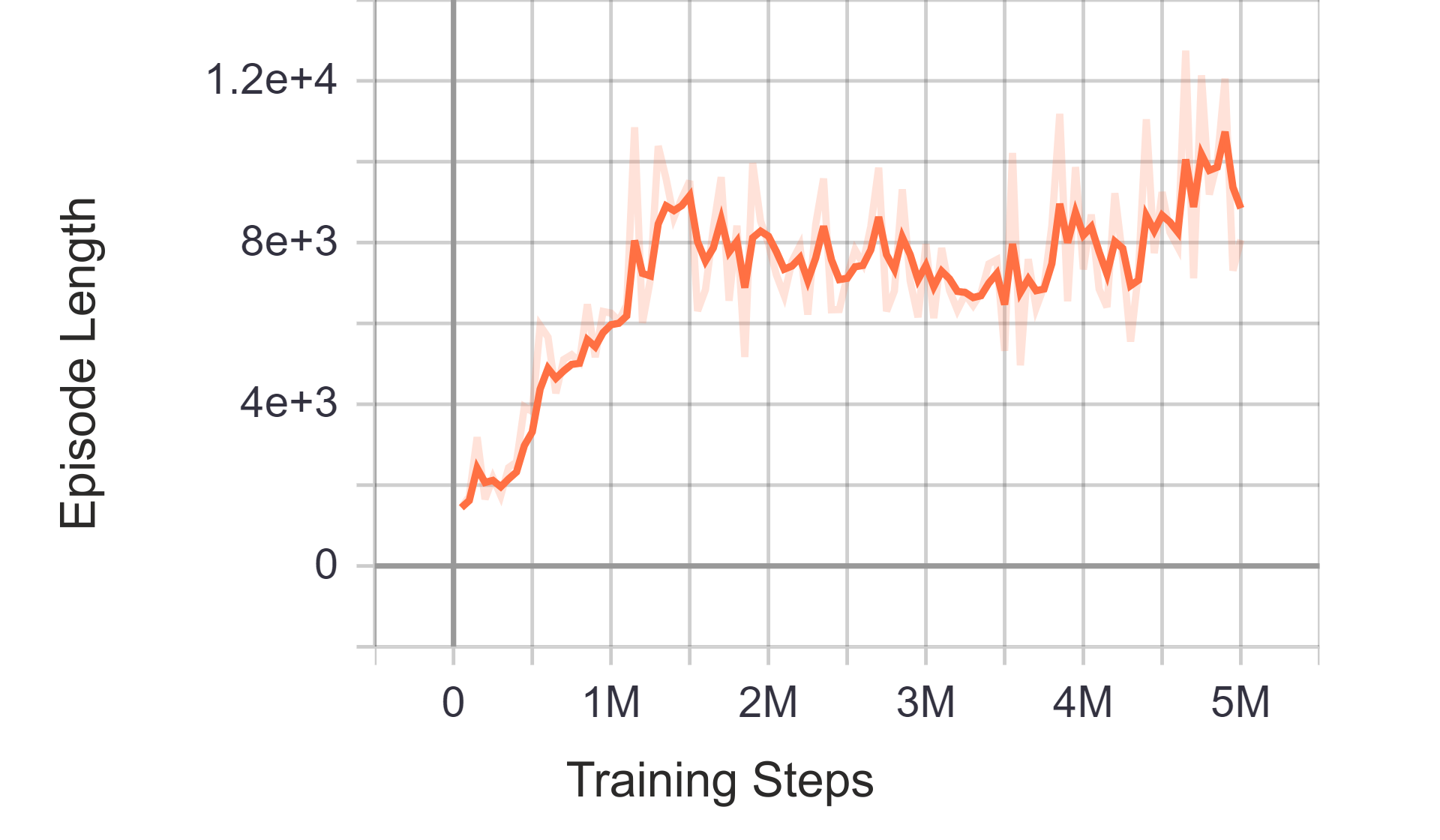}}
	\caption{Training results recorded for 5 million steps: (a) behavioral cloning loss, (b) GAIL reward, (c) curiosity reward, (d) extrinsic reward, (e) policy entropy, and (f) episode length. The occasional fluctuations in these statistics are a result of intermittent crashes while the agent was learning.}
	\label{fig4}
\end{figure}

The training phase of the proposed approach was analyzed in order to gain a better insight into the policy optimization process, and comment on the effectiveness of the hybrid learning strategy adopted in this work. Particularly, we analyzed the imitation learning (behavioral cloning loss, GAIL reward) and reinforcement learning (curiosity reward, extrinsic reward) metrics along with the policy entropy and episode length (refer Fig. \ref{fig4}).

Initially (until $\approx$ 0.5M training steps), the agent struggled to drive autonomously and often crashed into the roadside barriers at the first curve (ABC) itself; this was the principal reason for the episode length being too short ($\leq$ 4,000 decision steps) initially. Nevertheless, imitation learning components (behavioral cloning and GAIL reward) along with curiosity reward and collision penalty from reinforcement learning forced the agent to quickly update its policy, and it soon traversed the curve safely. This stage can be marked by the behavioral cloning loss reducing over 60\%, GAIL reward increasing up to 40 points, curiosity reward dropping nearly to zero, and extrinsic reward closing in on the 4,000 mark.

Up to this point, the agent had gathered enough experience to drive autonomously all the way up to the J$^{\text{th}}$ checkpoint. Here, the vehicle had attained high velocity owing to the prior straight drive (IJK), and the agent had to take a sharp left-right turn (JKL-KLM) using steering control alone (as the throttle and brakes of the vehicle were not controllable by the agent). This was a particularly critical maneuver, which was extremely difficult to master; over the next 0.5M training steps, the curiosity reward gradually increased a few points and the policy entropy (i.e., randomness) stagnated at 0.22. However, the agent eventually (after $\approx$ 1.5M training steps) learned to complete an entire lap around the racetrack, which caused the curiosity reward to fall back to zero and the extrinsic reward to shoot over 7,000 points. At this point in time, the behavioral cloning loss had almost converged to its minima ($\approx$ 0.6), indicating that imitation learning would not affect the policy parameters directly.

After completing its first autonomous lap, the agent seldom faced any serious issues while traversing the circuit in the following episodes (a strong effect of imitation learning). Although the vehicle continued to crash every once in a while, the cause of such crashes was mostly associated with the agent trying to optimize its trajectory even better. The agent soon approached closer to the global optima by enhancing its performance via reinforcement learning (2.5M to 4.5M steps) and started exploiting the learned behavior in order to optimize the policy in a closer neighborhood and minimize the randomness associated with mapping a specific observation vector to an appropriate action. Towards end of the training (4.5M to 5.0M steps), the agent settled within the neighborhood of its optimal performance (maximum GAIL reward of nearly 100 points and extrinsic reward of over 8,000 points), with a policy entropy of 0.16.

\subsection{Deployment Analysis}
\label{subsection: Deployment Analysis}

\begin{figure}[h]
	\centering
	\includegraphics[width=\linewidth]{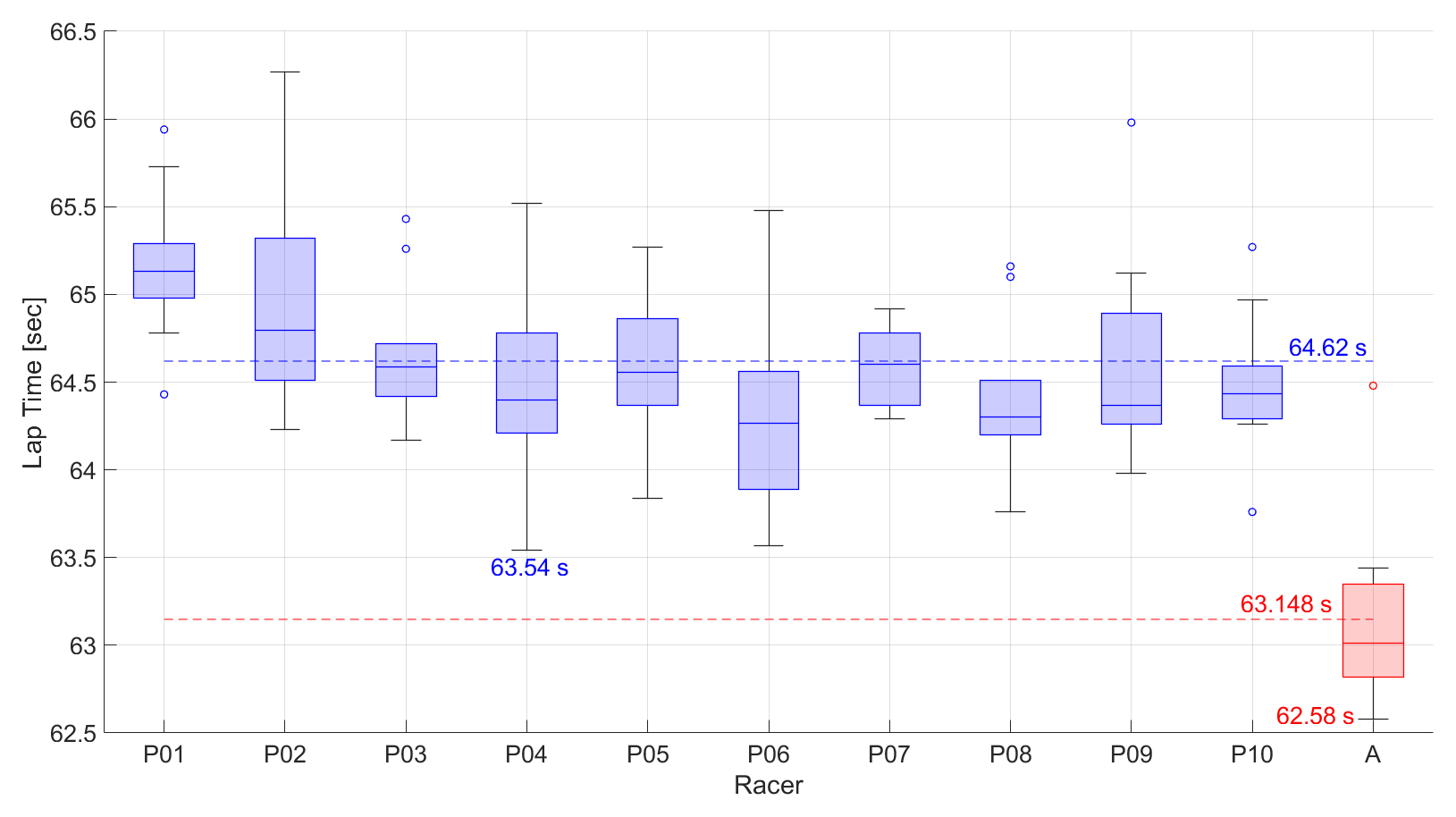}
	\caption{Lap time records of the autonomous agent $\left \{ \text{A} \right \}$ (represented in red color) compared on a common scale against those of 10 human players $\left \{ \text{P01}, \text{P02}, \cdots, \text{P10} \right \}$ (represented in blue color). The red and blue dashed lines indicate the mean lap times of autonomous agent (63.148 s) and human players (64.62 s) respectively. Also note the best lap time records of the autonomous agent (62.58 s) and human players (63.54 s).}
	\label{fig5}
\end{figure}

As discussed earlier, we compared the performance of our autonomous agent against 10 different human players in a virtual time attack car racing event. Fig. \ref{fig5} summarizes the lap time records of the said racing experiment. It is evident that the autonomous agent outperformed all the human players in most of the laps (9 out of 10), and with a significant difference: the autonomous agent was, on average, 1.46 seconds faster than the human players. Moreover, the autonomous agent defeated the best human player by a difference of 0.96 seconds\footnote{\textbf{Video:} \href{https://youtu.be/DMTuU2_NeUg}{https://youtu.be/DMTuU2\_NeUg}}, which is a significant lead compared to the close margins among top racers in similar racing events.

\begin{figure*}[t]
	\centering
	\subfigure[]{\includegraphics[width=0.49\textwidth]{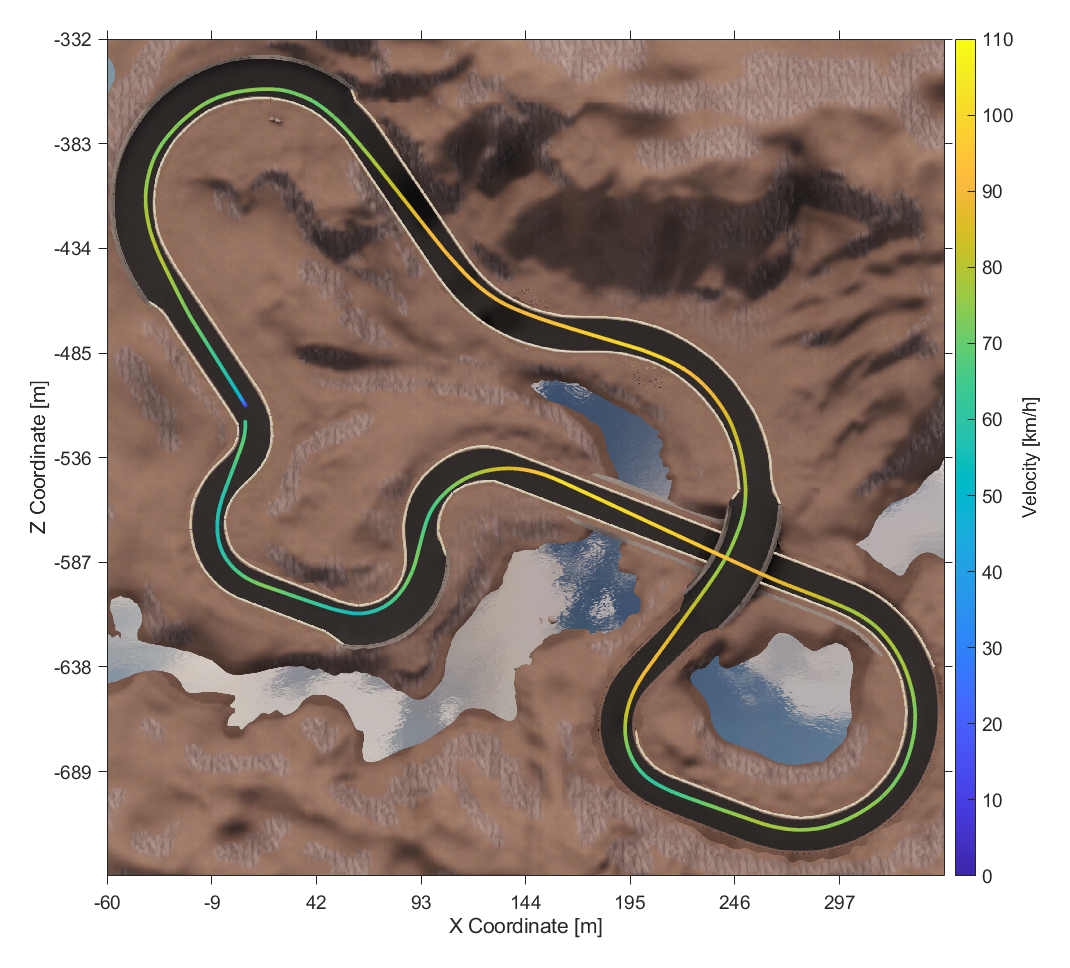}}
	\subfigure[]{\includegraphics[width=0.49\textwidth]{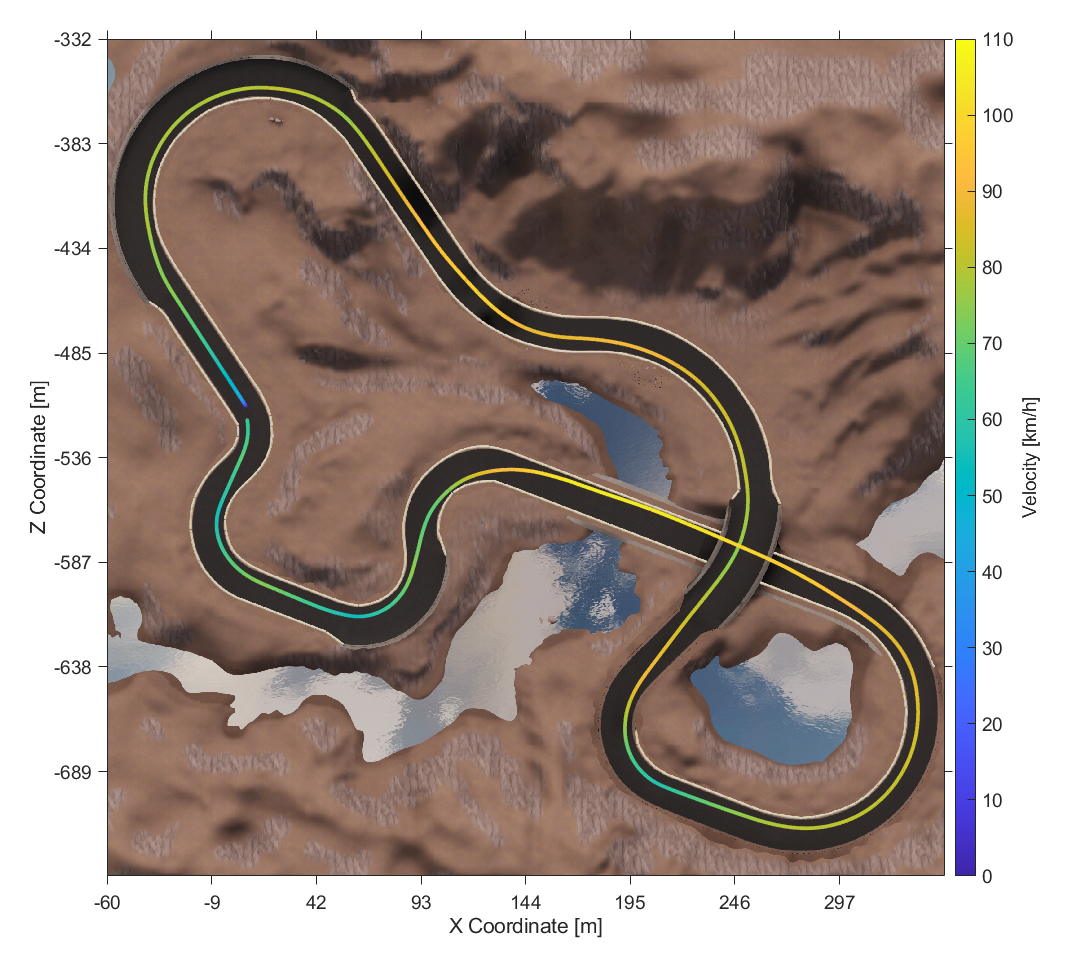}}
	\caption{Trajectory plots for the best (a) manual, and (b) autonomous laps recorded in a virtual time attack car racing event. The line plot indicates vehicle position over time and gradient of the trajectory indicates the instantaneous forward velocity of the vehicle at that position corresponding to each time step.}
	\label{fig6}
\end{figure*}

In order to gain a better insight into the racing experiment, and qualitatively compare the vehicle control strategy adopted by the autonomous agent with that of an expert human player, we analyzed the best manual and autonomous laps in terms of trajectories followed by both the entities, their actuation commands and their velocity profiles. It is to be noted that these results are obtained from one sample experiment and that the human player as well as the autonomous agent are extremely likely to follow a slightly different racing strategy in subsequent experiments owing to the fact that both the racer kinds are performing based on their ``learning'' and ``experience'' rather than explicitly hard-coded ``if-then-else'' rules.

Fig. \ref{fig6} depicts the trajectory plots for the fastest manual and autonomous laps. Upon taking a closer look, it can be observed that both the racers followed similar paths, except at sections CDE and IJK. Initially, owing to its well-rounded intelligence and experience in terms of virtual car racing, the human player slightly dominated the autonomous agent, and even performed a clean out-in-out maneuver at section CDE where the autonomous agent preferred a slightly curved path respecting the roadside barriers. It was truly a neck-and-neck competition until section IJK, which, as discussed earlier, was a specifically difficult maneuver for the autonomous agent as well as the human players. It is evident that even the best human racer was not able to handle this portion of the track very well; although the trajectory was collision-free, it was clearly sub-optimal. On the other hand, the autonomous racer had a significant advantage in terms of control precision, owing to which, it executed an excellent out-in-out maneuver (which is nearly unachievable by humans) over the said section. It is worth mentioning that the autonomous agent took a significant lead over the human racer particularly in this portion of the track.

\begin{figure*}[t]
	\centering
	\subfigure[]{\includegraphics[width=0.49\textwidth]{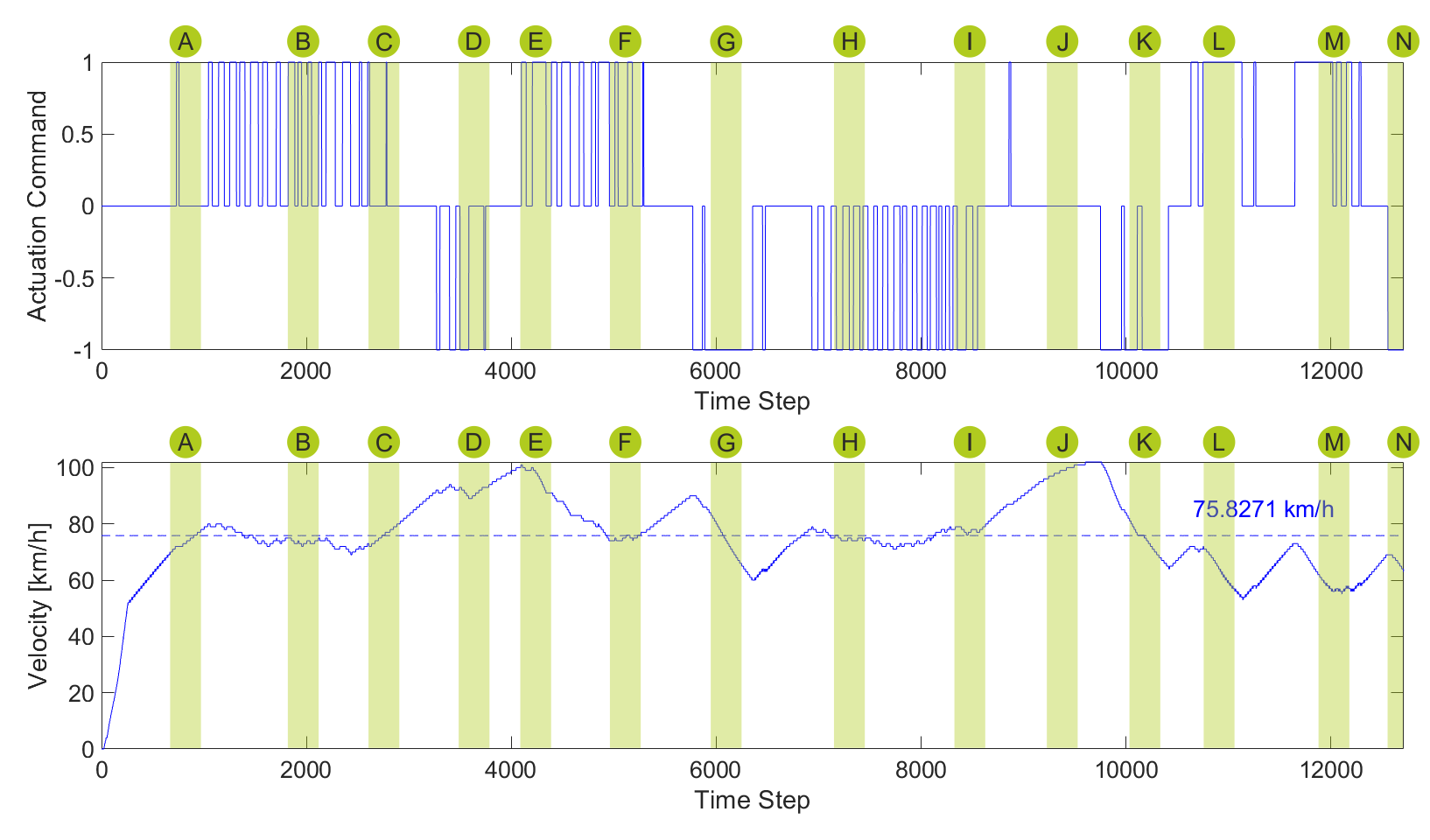}}
	\subfigure[]{\includegraphics[width=0.49\textwidth]{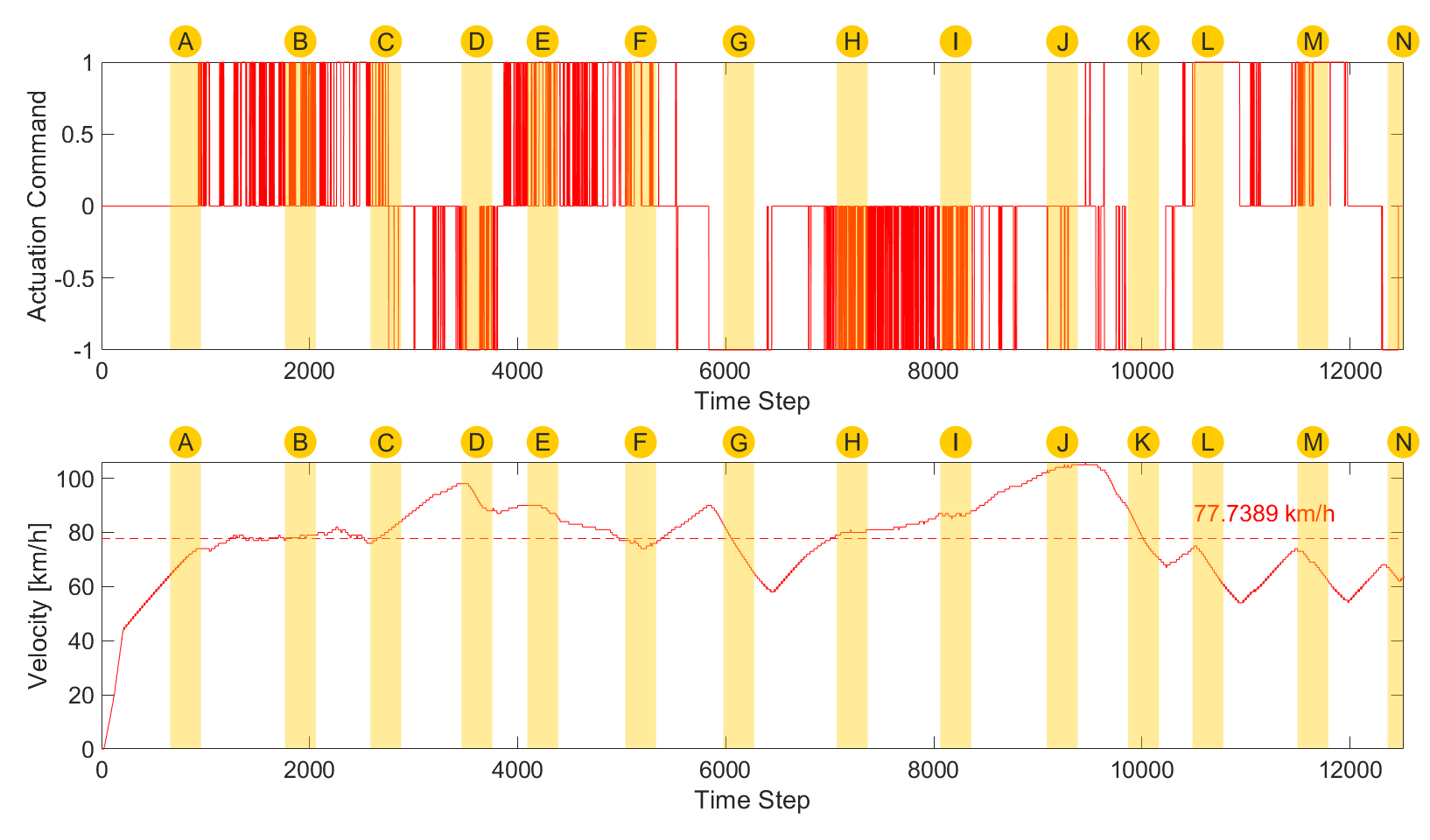}}
	\caption{Actuation commands and velocity profiles for the best (a) manual, and (b) autonomous laps recorded in a virtual time attack car racing event. The dashed lines in velocity profiles indicate the average velocity of the human player (75.83 km/h) and autonomous agent (77.74 km/h) respectively.}
	\label{fig7}
\end{figure*}

Apart from trajectory optimization, another important strategy adopted by the autonomous agent so as to consistently maximize the vehicle speed was to avoid continuous drifting over cornering maneuvers by frequently switching the steering command. This allowed the agent to align the vehicle wheels in the direction of velocity vector so as to minimize drifting, while effectively achieving the required vehicle orientation as well. As discussed earlier, most autonomous systems have an upper hand over humans when it comes to control resolution and switching frequency. Since human response time in terms of providing control command to the vehicle (using standard computer keyboard as the input source) was quite low as compared to that of the autonomous agent, the time difference in action switching was significantly high (refer Fig. \ref{fig7}). This was the key phenomenon that restrained the human player from outperforming the autonomous agent despite its superior trajectory optimization in the initial portion of the track.

Finally, we comment on the ultimate parameter of interest for any racing event -- vehicle velocity (refer Fig. \ref{fig7}). As discussed earlier, since the human player optimized its trajectory better in the initial portion of the track, its velocity was slightly higher (in some instances) than the autonomous agent. Nevertheless, the difference was negligible as the autonomous agent too optimized its velocity by minimizing drifting. In fact, this ability of the autonomous agent, to avoid continuous drifting, not only mitigated the velocity difference in the earlier phase of the racing event, but it also contributed significantly towards the overall velocity-oriented dominance that the autonomous agent exhibited over the human player for the remainder portion of the lap. Particularly, after section CDE (i.e., after $\approx$ 4,200 time steps w.r.t. Fig. \ref{fig7}), it can be observed that the instantaneous forward velocity of the vehicle in case of autonomous lap was almost always higher than that of the manual lap. As a result, the average (statistical mean taken over all the data points recorded from one complete lap) velocity of the autonomous agent (77.74 km/h) was significantly higher than that of the human racer (75.83 km/h).

\subsection{Computational Analysis}
\label{subsection: Computational Analysis}

We made use of a personal computer incorporating Intel i7-8750H CPU, NVIDIA RTX 2070 GPU and 16 GB RAM to train and deploy the autonomous racing behavior model using the AutoRACE Simulator. Specifically, the simulator application was executed atop Unity 2018.4.24f1 and the machine learning task was handled by ML-Agents 0.19.0 in conjunction with Python 3.7.9 and TensorFlow 2.3.0. While the simulator itself exploited both CPU and GPU resources for its backend and frontend operations respectively, the neural network training and inference phases were both carried out on the CPU alone. It is also to be noted that all the temporal measurements presented within this section were recorded while running the simulator and training/deployment scripts in parallel on a single laptop with the aforementioned configuration.

Training the autonomous racing behavior for a single agent using the proposed hybrid imitation-reinforcement learning architecture for 5 million steps required 19 hours 39 minutes and 51 seconds. It is worth mentioning that compared to the latest approach aimed at super-human autonomous racing \cite{fuchs2020}, which took over 72 hours of training time with four PlayStation 4 consoles simulating 20 cars each and a dedicated workstation training the neural network policy on its GPU, our approach trained the autonomous agent much faster despite the limitation in terms of computational hardware. The hybrid architecture greatly aided in reducing the training time since the collision rate of the vehicle quickly reduced over time (as a result of autonomous driving ability learned through imitation) and the custom extrinsic reward function efficiently guided the agent towards its ultimate objective of minimizing lap time.

The mean deployment latency of the proposed approach was 3.42 milliseconds for a single observation-action cycle, while a complete simulation step including other computational components such as rendering, scripts, physics, transformations, illumination, user interface, etc. took about 9.61 milliseconds, on average. Note that such a low latency was achievable owing to the end-to-end nature of the neural network policy. It can be invariably argued that this low latency algorithm allowed the autonomous agent to control the vehicle at a much higher rate, potentially offering it a significant advantage over the expert human competitors, as discussed earlier.


\section{Conclusion}
\label{section: Conclusion}

This work introduced AutoRACE Simulator, an open, flexible and realistic dynamic simulation system aimed at accelerated development, deployment and analysis of autonomous racing algorithms. The said simulator can be employed to implement both modular as well as end-to-end racing solutions, and offers a manual override switch to compare autonomous racing algorithms against human expertise. The simulator also features an automated data logging support to record racing events for further analysis.

In this work, we proposed and adopted a hybrid imitation-reinforcement learning architecture along with a novel reward function to train an autonomous agent to race aggressively in less than 20 hours. While the agent learned to drive using imitation learning, it enhanced its performance recursively through reinforcement learning. Imitation learning also restricted the agent from cheating through unethical actions, which significantly contributed towards a faster and safer training process. The agent ultimately learned the objective of minimizing lap times around a virtual circuit by optimizing its trajectory and avoiding continuous drifting at cornering maneuvers.

We analyzed the trained neural network model by comparing the performance of our autonomous agent against that of 10 expert human players in a simulated time attack car racing event. Results indicated that the autonomous agent was on average, 1.46 seconds faster than the human players and strongly dominated them. A direct comparison of the best autonomous and manual laps further supported this claim, wherein the autonomous agent defeated the best human racer by a difference of 0.96 seconds. This win could be accounted partly by the perfection of the autonomous agent in terms of trajectory optimization and partly by its higher control frequency, which allowed it to minimize unwanted drifting.

This work can be extended to other race forms with distinct rules, vehicle models, environmental settings, agent semantics and actor counts. Additionally, effect of variations in observation and action spaces on the agent's performance can be evaluated exhaustively. Furthermore, this work can be taken up to transfer the racing behavior models trained in simulated scenarios to real-world autonomous racecars, with little or no modification. Finally, this platform can be exploited to benchmark autonomous racing solutions by recording and tracking performance metrics of different human players and autonomous agents.


\balance
\bibliographystyle{IEEEtran}
\bibliography{Bibliography}

\end{document}